%% file: xtroll.tex
\documentclass[sigconf]{acmart}
\AtBeginDocument{%
  }

\copyrightyear{2025}
\acmYear{2025}
\setcopyright{rightsretained}
\acmConference[CIKM '25] {Proceedings of the 34th ACM International Conference on Information and Knowledge Management}{ November 10--14, 2025}{Seoul, Republic of Korea.}
\acmBooktitle{Proceedings of the 34th ACM International Conference on Information and Knowledge Management (CIKM '25), November 10--14, 2025, Seoul, Republic of Korea}
\acmISBN{979-8-4007-2040-6/2025/11}
\acmDOI{10.1145/XXXXXX.XXXXXX}

\usepackage{times}
\usepackage{url}
\usepackage{amsthm}
\usepackage{booktabs}
\usepackage{algorithm}
\usepackage{booktabs}
\usepackage{comment}
\usepackage{subfig}
\usepackage{fancyhdr}
\usepackage{amsmath}
\usepackage{xurl}
\usepackage{adjustbox}
\usepackage{arydshln}
\usepackage{relsize}
\usepackage{wasysym}
\usepackage{multirow}
\usepackage[makeroom]{cancel}
\usepackage{scalerel,graphicx,xparse}
\usepackage{xcolor}
\usepackage{algpseudocode}
\usepackage{enumitem}
\usepackage[most]{tcolorbox}

\usepackage{lscape}
\usepackage[capitalize,nameinlink]{cleveref}

\begin{document}

\title{X-Troll: eXplainable Detection of State-Sponsored Information Operations Agents}


\author{Lin Tian}
\email{Lin.Tian-3@uts.edu.au}
\affiliation{%
  \institution{University of Technology Sydney}
  \city{Sydney}
  \country{Australia}
}

\author{Xiuzhen Zhang}
\email{xiuzhen.zhang@rmit.edu.au}
\affiliation{%
  \institution{RMIT University}
  \city{Melbourne}
  \country{Australia}
}

\author{Maria Myung-Hee Kim}
\email{myung.kim@defence.gov.au}
\affiliation{%
  \institution{Defence Science and Technology Group}
  \country{Australia}
}

\author{Jennifer Biggs}
\email{jennifer.biggs@defence.gov.au}
\affiliation{%
  \institution{Defence Science and Technology Group}
  \country{Australia}
}

\author{Marian-Andrei Rizoiu}
\email{Marian-Andrei.Rizoiu@uts.edu.au}
\affiliation{%
  \institution{University of Technology Sydney}
  \city{Sydney}
  \country{Australia}
}



\begin{abstract}
State-sponsored trolls, malicious actors who deploy sophisticated linguistic manipulation in coordinated information campaigns, posing threats to online discourse integrity. While Large Language Models (LLMs) achieve strong performance on general natural language processing (NLP) tasks, they struggle with subtle propaganda detection and operate as ``black boxes'', providing no interpretable insights into manipulation strategies.
This paper introduces \textbf{X-Troll}, a novel framework that bridges this gap by integrating explainable adapter-based LLMs with expert-derived linguistic knowledge to detect state-sponsored trolls and provide human-readable explanations for its decisions.
X-Troll incorporates appraisal theory and propaganda analysis through specialized LoRA adapters, using dynamic gating to capture campaign-specific discourse patterns in coordinated information operations.
Experiments on real-world data demonstrate that our linguistically-informed approach shows strong performance compared with both general LLM baselines and existing troll detection models in accuracy while providing enhanced transparency through expert-grounded explanations that reveal the specific linguistic strategies used by state-sponsored actors.
X-Troll source code is available at: \url{https://github.com/ltian678/xtroll_source/}.
\end{abstract}




\maketitle

\section{Introduction}
State-sponsored information operation agents—commonly known as troll accounts—have emerged as major threat actors in the digital information ecosystem, systematically manipulating public discourse to achieve geopolitical objectives~\cite{vosoughi2018spread,lazer2018science}. Unlike isolated bad actors spreading misinformation, these agents operate as coordinated units within state-directed campaigns, using nuanced linguistic strategies and assuming false personas to infiltrate and influence online communities. Their tactics extend beyond spreading falsehoods to include more subtle forms of manipulation: amplifying divisive content, undermining institutional trust, and steering narratives through strategic emotional appeals.
A prominent example is the \texttt{Doppelg\"{a}nger campaign}, a coordinated disinformation effort that mimicked legitimate media sources to spread misleading narratives\footnote{\url{https://www.disinfo.eu/doppelganger-operation/}}.
Volunteer groups like \texttt{@antibot4navalny} have played a crucial role in exposing these operations by manually tracking and documenting troll activities, providing valuable insights into the evolving nature of disinformation. However, the scale of these campaigns substantially exceeds the capacity of manual efforts, highlighting the necessity for automated and scalable troll detection.

Automated disinformation and troll detection have been widely explored 
in the literature~\cite{tian2023metatroll,addawood2019linguistic,im2020still,shafiei2022detection,dutt2018senator,badawy2019falls}. 
In particular, neural approaches, such as pre-trained language models and graph neural networks, have been used to analyze the content of social media posts and their propagation patterns, and user engagement patterns for detecting troll behavior. 
However, most existing models 
operate as black-boxes, offering little transparency into how decisions are made. 
This lack of interpretability limits their real-world usability, as end-users often struggle to trust automated predictions. 
Research on machine learning interpretability has evolved from explaining internal decision-making processes to generating human-understandable explanations, particularly in NLP applications.
Rationalization, in particular, involves identifying input phrases
sufficient to predict the desired outcome~\cite{lei2016rationalizing,liu2022fr}.
While most studies on rationalization primarily focus on generating phrase-level explanations, the challenge of producing natural language justifications that are easily understandable to humans remains largely unsolved.

Media and communication research
provides critical insights into disinformation by exploring its creation, spread, and impact across media platforms. 
Experts in these fields have identified distinctive 
linguistic features of disinformation discourse \cite{martin2003language,inwood2021ambient}, the mechanisms of its dissemination and amplification \cite{zhang2021assembling,moral2023overview}, and the psychological and societal effects on audiences \cite{inwood2021ambient}. 
Such studies have shown that trolls employ propagandistic and targeted communication strategies, often characterized by inflammatory and provocative language, to influence audiences~\cite{moral2023overview}. 
These human-derived expert insights provide deep contextual understanding that complements computational machine learning research.

To address the limitations revealed by recent LLM evaluations and leverage the insights from linguistic analysis, we pose two key research questions:

\noindent\textbf{RQ1:} Why do transformer-based models trained on massive corpora struggle with detecting state-sponsored trolls despite strong performance on related tasks? What linguistic knowledge—specifically from appraisal theory and propaganda analysis—is required to identify coordinated manipulation patterns?

\noindent\textbf{RQ2:} How can we generate explanations that reveal not just what features triggered a classification, but the underlying manipulation strategies being used, when dealing with adversaries who deliberately obscure their tactics?

To answer these research questions, we propose X-Troll, a framework that integrates linguistic expert knowledge -- specifically appraisal analysis and propaganda strategy identification -- through Low-Rank Adaptation (LoRA) fine-tuning of LLMs. Unlike general-purpose models that struggle with subtle propaganda techniques, X-Troll analyzes users' social media post timelines through the lens of established linguistic theory to perform classification while producing expert-informed rationales. By combining domain-specific linguistic knowledge with adapter-based fine-tuning and rationale-based explanation generation, X-Troll is trying to mitigate the limitations of general LLMs in propaganda detection while achieving high accuracy and producing human-readable natural language explanations that illuminate the specific linguistic manipulation strategies used by state-sponsored trolls.

\section{Related Work}
State-sponsored troll and disinformation campaign detection has become a key research focus. 
Recent work primarily uses NLP techniques for Information Campaigns.
\citet{Kim2019} developed a time-sensitive semantic edit distance (t-SED) metric to analyze user identity and social roles through timestamped text sequences. 
Their case study of Russian trolls on Twitter classified social roles into left-, right-leaning, and news feed categories.
\citet{addawood2019linguistic} explored linguistic cues that indicate deception in political trolls' social media posts. They identified key markers of state-sponsored accounts. \citet{im2020still} proposed a content-based approach to detect Russian troll accounts on Twitter. Their method leverages user metadata, activity patterns, and linguistic features. 

Behavioral modeling approaches have shown promise in capturing troll dynamics.
\citet{Rizoiu2018a} found that socialbots are 2.5 times more influential than humans during political events, while \citet{Ram2021a} developed birdspotter, an end-to-end pipeline achieving state-of-the-art bot detection performance. 
Recent advances focus on early detection and social system reactions. 
\citet{Tian2025ICMamba} introduced IC-Mamba for predicting engagement patterns within 15-30 minutes of posting, while \citet{Kong2023} proposed detecting information operations by analyzing social reactions and retweeting patterns to identify state-backed agents. 
\citet{Ram2025} provided ideology detection guidelines, showing that right-wing ideologies exhibit distinct patterns in moral language and dichotomous thinking.

Understanding the broader manipulation ecosystem provides crucial context for detection systems. 
\citet{Calderon2024a} introduced the Opinion Market Model to evaluate interventions against extremist content spread, while \citet{Kong2022} used mixed methods to explain how extreme opinions infiltrate mainstream discussions. 
Crisis events amplify these dynamics: \citet{Bailo2023} found far-right accounts moved from peripheral to central positions during Australian disasters, and \citet{Johns2024} showed that some Facebook pages overperformed during COVID-19 despite content moderation efforts. 
\citet{ferrara2020characterizing} surveyed the landscape of social bots and their role in manipulation campaigns, while \citet{zannettou2019disinformation} traced how disinformation spreads across the web ecosystem. \citet{starbird2019disinformation} revealed how alternative media ecosystems participate in information operations, highlighting the need for comprehensive detection approaches. These studies underscore that troll detection cannot be isolated from understanding the broader information context.

Explainability in rumor and fake news detection has gained increasing attention. 
Early efforts primarily employ attention mechanisms to generate explanations. 
These methods leverage both text and non-text signals to provide insights into detection model decisions~\cite{shu2019defend,khoo2020interpretable,ni2021mvan,lu2020gcan,silva2021propagation2vec,liu2023towards}.
For example, \citet{shu2019defend} applied co-attention mechanisms to examine the relationship between content and audience reactions. 
This helps classify news as real or fake. 
Similarly, \citet{khoo2020interpretable} used multi-head attention to analyze tweet interactions at both token and post levels.
However, attention mechanisms are not inherently designed for human interpretability.

Multi-task learning approaches have shown promise in related domains. 
\citet{Yuan2023,Yuan2024} demonstrated that training simultaneously across multiple hate speech datasets significantly improves generalization to unseen domains, achieving substantial improvements over single-task approaches. 
Their multi-task learning pipeline parallels our multi-adapter architecture, where each adapter specializes in different aspects of manipulative discourse while sharing the base model representation.

Some studies have utilized user attributes and propagation patterns to explain detection model decisions. 
\citet{vosoughi2018spread} identified distinct propagation patterns for fake news. 
\citet{ni2021mvan} extended this work by modeling propagation with graph neural networks. 
They used attention mechanisms to highlight key features.
\citet{lu2020gcan} and \citet{silva2021propagation2vec} further explored these patterns to distinguish fake news from true news. 
Both employed attention mechanisms to provide explanations. 
\citet{liu2023towards} advanced this work using self-supervised graph learning. 
Their approach identifies important nodes and generates interpretable subgraphs.

Interpretability remains a challenge in AI and machine learning, particularly for large language models (LLMs). 
Recent research has focused on the plausibility of model-generated rationalizations.
\citet{rajani2019explain} introduced Commonsense Auto-Generated Explanations. 
Their approach fine-tunes language models on explanation datasets. 
This automatically generates rationalizations for commonsense QA tasks.
\citet{liuetal2023mgr} extended this work by developing a model with multiple generators and a single predictor. 
Each generator uses different initializations to produce diverse rationale candidates. 
This design reduces bias and improves explanation plausibility.
While these developments show promise, their effectiveness for troll detection remains unclear.

Despite progress in automated troll detection and broader machine learning interpretability, explainability of troll detection systems remains largely unexplored. 
This paper addresses this gap by proposing a novel rationale-based approach to troll detection.
We aim to provide clear and human-understandable explanations for detection model decisions.

\section{Preliminaries}
In this section, we provide the necessary background on appraisal theory and propaganda analysis.
Effective troll detection requires understanding the systematic linguistic strategies that distinguish coordinated manipulation from authentic discourse. We ground X-Troll in established theoretical frameworks from discourse analysis and propaganda studies, enabling both accurate detection and interpretable explanations of manipulative communication patterns.


\subsection{Appraisal Theory} 
Appraisal theory~\cite{martin2003language} provides a systematic framework for analyzing how language users express evaluative stance and emotional positioning in discourse. Recent research has demonstrated that state-sponsored trolls exhibit distinctive appraisal patterns that differ systematically from authentic users \cite{tian2023task}, making this framework particularly valuable for troll detection.

\noindent\textbf{Theoretical Framework.} Appraisal analysis examines three interconnected systems of evaluative meaning: Attitude (emotional reactions and judgments), Engagement (how speakers position themselves relative to their propositions), and Graduation (the scaling of evaluative intensity). For troll detection, we focus on three dimensions that capture manipulative discourse strategies:

\textit{Ideational targeting:} Systematic focus on specific entities, topics, or themes designed to direct audience attention toward predetermined narrative frames. State-sponsored trolls consistently target particular political figures, institutions, or ideological concepts rather than engaging in organic topical variation.

\textit{Sentiment polarity:} Strategic deployment of positive, negative, or neutral evaluative language to influence audience perception. Unlike authentic users who express emotional responses, trolls systematically calibrate sentiment to achieve specific persuasive objectives.

\textit{Persona construction:} Linguistic techniques used to establish false credibility or deliberately provoke emotional responses. This includes strategic deployment of authority markers, community membership signals, and emotional authenticity performance.

\noindent\textbf{Empirical Application.} We operationalize appraisal analysis through expert annotation of state-sponsored troll datasets, focusing on Twitter--released accounts from Russian Internet Research Agency operations (detailed in \cref{sec:experiments}). Domain experts identified systematic patterns where trolls manipulate evaluative language to maximize influence. These patterns provide supervised signals that enable X-Troll to recognize subtle linguistic manipulation strategies that automated systems typically miss.


\subsection{Propaganda Technique Analysis} 
State-sponsored trolls systematically deploy specific propaganda techniques to manipulate audience perception and achieve strategic objectives. 

\noindent\textbf{Theoretical Foundation.} We integrate propaganda technique identification to capture these explicit manipulation strategies, building on established taxonomies and modern computational approaches~\cite{lee1939fine,da2019fine}.
We focus on three core techniques commonly used in state-sponsored information operations:

\textit{Loaded Language} involves strategic use of emotionally charged terms to provoke reactions rather than facilitate rational discourse, by associating targets with predetermined emotional valences.

\textit{Appeal to Commonality} frames partisan positions as widely accepted beliefs, creating false consensus through linguistic manipulation that exploits social proof heuristics.

\textit{Doubt and Questioning} systematically undermines credible sources and established facts through persistent skepticism, eroding confidence in existing knowledge structures without providing alternative explanations.

\noindent\textbf{Integration with X-Troll.} We incorporate the DIPROMATS 2023 dataset~\cite{moral2023overview}, which provides post-level annotations of these techniques in real-world information operations. These annotations serve as supervised signals enabling X-Troll to recognize established propaganda strategies across user timelines. This propaganda-aware approach provides two advantages: distinguishing coordinated operations from authentic communication, and supplying structured vocabulary for interpretable explanations grounded in established rhetorical analysis.

\begin{figure*}[t]
\centering
  \includegraphics[width=\linewidth]{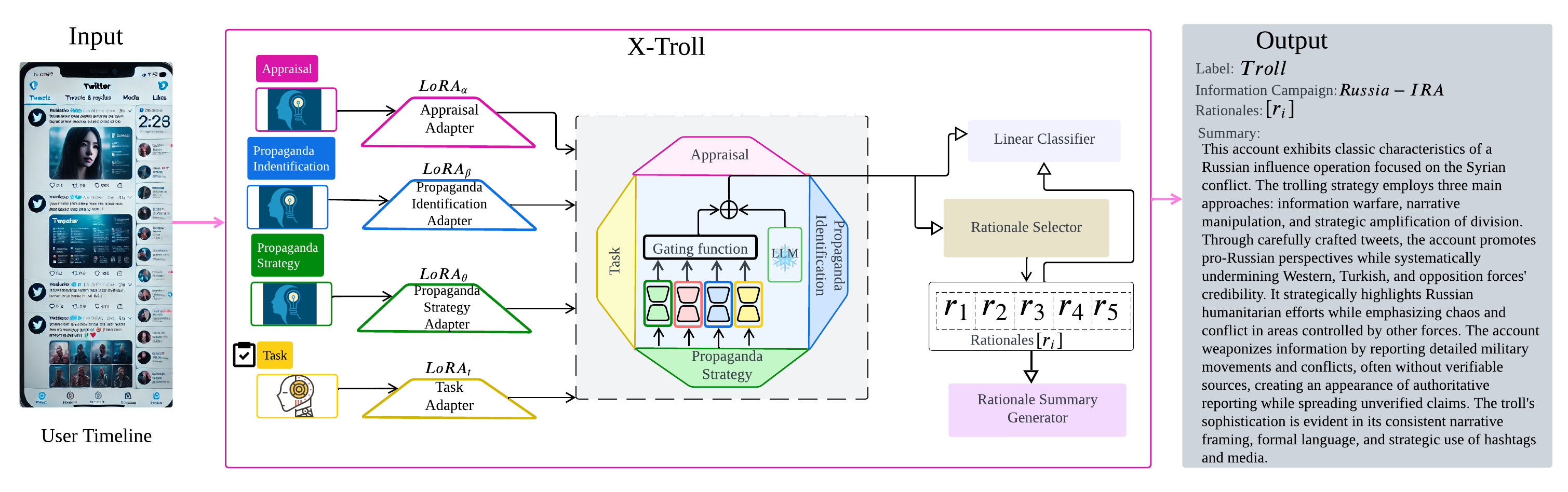}
  \caption{X-Troll framework for explainable state-sponsored troll detection. Given a user timeline, four LoRA adapters capture distinct aspects of manipulative discourse: Appraisal (evaluative language patterns), Propaganda Identification (binary propaganda detection), Propaganda Strategy (specific manipulation techniques), and Task (troll-specific features). A dynamic gating mechanism adaptively weights adapter contributions, feeding the fused representation to a linear classifier for troll detection and campaign classification. The rationale selector identifies salient tokens across the timeline, which the summary generator transforms into human-readable explanations grounded in linguistic theory. The example shows detection of a Russian-IRA information operation with extracted rationales and generated explanation revealing narrative manipulation strategies.}
  \label{fig:overall}
\end{figure*}




\section{Methodology}
\label{sec:methodology}
This section introduces our problem formulation and presents X-Troll, a rationale-based framework for explainable detection of state-sponsored trolls. As shown in \cref{fig:overall}, X-Troll integrates three core components: (1) knowledge-fused LLM adapters for multifaceted feature extraction from user timelines, (2) a unified rationale selector for identifying key trolling evidence, and (3) a summary generator for producing human-readable explanations from selected rationales.

\subsection{Problem Statement}
\label{sec:problem_statement}
Social media troll detection differs from general misinformation detection. 
While misinformation detection focuses on post-level falsehood, troll accounts engage in information campaigns with specific interests and targets over time. 
Troll user timelines exhibit distinct linguistic and behavioral features that differ from typical social media users.

We formalize the problem as follows.
Given a set of users $\mathcal{U} = \{u_1, \ldots, u_m\}$ and a set of information campaigns $\mathcal{C}$, the task is to detect troll users and identify their associated information campaigns. 
Each user $u$ has a timeline of posts $T_u = [x_{(u,1)}, \ldots, x_{(u,n)}]$, where $x_{(u, j)}$ represents the $j$-th post by $u$. Users are labeled as \textit{trolls} ($y_u = 1$) or \textit{non-trolls} ($y_u = 0$).

\noindent\textit{Troll and Information Campaign Classification.} 
A classifier $g_\phi$ predicts whether a user is a troll based on their timeline: $y_u = g_\phi(T_u)$. For users identified as trolls, an additional classifier $h_\theta$ predicts their associated information campaign $c_u \in \mathcal{C}$: $c_u = h_\theta(T_u)$.

\noindent\textit{Rationale Selection and Explanation Generation.} 
We address explainability through a two-stage interpretability mechanism. 
First, a rationale selector $f_\gamma$ identifies a sparse subset of $k$ salient tokens across the user's timeline. 
These tokens most strongly influence the classification decision: $\mathcal{R}_u = f_\gamma(T_u, y_u)$. 
Second, an explanation generator $s_\omega$ synthesizes a comprehensive, human-readable explanation from the extracted rationales: $S_u = s_\omega(\mathcal{R}_u)$. 

\subsection{Timeline Encoding}
To model user behaviour over time, we encode a user's timeline $T_u$ consisting of posts $\{x_{(u,j)}\}_{j=1}^{n_u}$. Each post is first transformed into a contextualized representation using a pretrained language model $f_{\beta}$: $\mathbf{h}_{(u,j)} = f_{\beta}(x_{(u,j)}) \in \mathbb{R}^d$. 
Next, the sequence of post embeddings $[\mathbf{h}_{u,1}, \ldots, \mathbf{h}_{u,n_u}]$ is processed by a Transformer encoder parameterized by $\gamma$: 
$\mathbf{H}_u = \text{Transformer}_{\gamma}([\mathbf{h}_{(u,1)}, \ldots, \mathbf{h}_{(u,n_u)}]) \in \mathbb{R}^{n_u \times d}$,
where  $\mathbf{H}_u = [\tilde{\mathbf{h}}_{(u,1)}, \ldots, \tilde{\mathbf{h}}_{(u,n_u)}]$  represents the enhanced timeline representations. 
To obtain a single timeline representation \(\mathbf{t}_u\), we apply attention pooling, which assigns importance weights to individual posts: \begin{align*}
\alpha_{u,j} &= \frac{\exp(\mathbf{q}^\top \tilde{\mathbf{h}}_{(u,j)})}{\sum_{k=1}^{n_u} \exp(\mathbf{q}^\top \tilde{\mathbf{h}}_{(u,k)})}, \\
\mathbf{t}_u &= \sum_{j=1}^{n_u} \alpha_{(u,j)},\tilde{\mathbf{h}}_{(u,j)},
\end{align*} where \(\mathbf{q} \in \mathbb{R}^d\) is a learnable query vector. We experimented with mean pooling, max pooling, and attention-based pooling, finding that attention pooling outperformed other methods by emphasizing the most informative posts for downstream tasks.

\subsection{Adapter Fusion}
X-Troll uses a novel adapter fusion architecture based on Low-Rank Adaptation (LoRA)~\cite{hu2022lora} to incorporate diverse expert knowledge while maintaining computational efficiency. 
This approach enables parameter-efficient fine-tuning while integrating domain-specific knowledge across multiple dimensions of troll behavior. 
Each adapter specializes in a distinct aspect of troll behavior and is dynamically integrated via a \textit{gating mechanism} (\cref{fig:overall}).

\subsubsection{LoRA Adapter Mechanism}
We leverage LoRA to efficiently fine-tune X-Troll. 
LoRA adapts pretrained language models while keeping their weights frozen. 
Given a weight matrix $W \in \mathbb{R}^{d \times k}$, LoRA introduces low-rank matrices $B \in \mathbb{R}^{d \times r}$ and $A \in \mathbb{R}^{r \times k}$, where $r \ll \min(d,k)$, updating weights as: $W' = W + BA$. 
This enables task-specific knowledge adaptation while maintaining computational efficiency. 
Each adapter operates independently while sharing the base model, facilitating multi-domain adaptation. 
The LoRA update modifies the hidden state representation $h$: $h'  = Wh + \Delta Wh = Wh + B(Ah)$. 
This ensures domain-relevant transformations without full fine-tuning. 
X-Troll integrates three specialized LoRA-adapters designed to capture strategic, linguistic, and behavioral aspects of troll detection.

\paragraph{(1) Appraisal Adapter ($\text{LoRA}_\alpha$)} 
It performs fine-grained linguistic analysis based on appraisal theory, trained via token-level sequence labeling on expert-annotated data. It captures ideational targeting (consistent entity focus), sentiment polarity (strategic emotional framing), and persona construction (credibility establishment techniques). The adapter optimizes a custom sequence labeling loss $\mathcal{L}_\text{appraisal}$ accounting for hierarchical appraisal features across linguistic spans.

\paragraph{(2) Propaganda Identification Adapter ($\text{LoRA}_\beta$)} 
It specializes in detecting binary propaganda presence within posts, drawing from the DIPROMATS 2023 propaganda dataset annotations. It applies targeted low-rank updates ($\Delta \mathbf{W}_\beta = \mathbf{B}_\beta\mathbf{A}_\beta$) to model layers most sensitive to propaganda features. The adapter is optimised using binary cross-entropy loss $
\mathcal{L}_\text{prop} = -\frac{1}{N}\sum_{i=1}^{N}[y_i\log(p_i) + (1-y_i)\log(1-p_i)],$
where $y_i$ represents the ground-truth propaganda label and $p_i$ the model prediction for the $i$-th sample.

\paragraph{(3) Propaganda Strategy Adapter ($\text{LoRA}_\theta$):} This adapter performs fine-grained classification of specific propaganda techniques (e.g.,\ loaded language, appeal to fear, causal oversimplification) based on multi-class strategy annotations. It is optimised using a categorical cross-entropy loss $
\mathcal{L}_\text{strat} = -\frac{1}{N}\sum_{i=1}^{N}\sum_{c=1}^{C}y_{(i,c)}\log(p_{(i,c)}),$
where $C$ represents the number of propaganda strategy classes, and $y_{i,c}$ and $p_{i,c}$ represent the ground truth and prediction for class $c$ of sample $i$, respectively.

\paragraph{(4) Task Adapter ($\text{LoRA}_t$):} This adapter serves as the task-specific component directly optimised for troll detection and identification, capturing patterns distinctive to state-sponsored trolls. It is trained using the troll classification loss:
\begin{equation*}
\mathcal{L}_\text{task} = -\frac{1}{N}\sum_{i=1}^{N}[y_i\log(p_i) + (1-y_i)\log(1-p_i)].
\end{equation*}

By integrating expert knowledge through these specialised adapters, X-Troll captures patterns of troll behaviour while maintaining computational efficiency.
Our multi-adapter approach builds on behavioral pattern analysis findings. 
\citet{Yuan2025} demonstrated that behavioral homophily can reveal user patterns that transcend topical similarity, suggesting that similar behavioral policies may be identifiable across different topics. This supports our design of specialized adapters that capture different aspects of trolling activities.

\subsubsection{Dynamic Gating Mechanism}
To effectively integrate outputs from multiple specialised LoRA adapters, X-Troll uses a dynamic gating mechanism that adaptively learns the optimal weighting of each adapter's contribution. Let $K$ denote the number of adapters; in our case, $K = 4$.

Each adapter produces an output representation $h_k \in \mathbb{R}^d$,  
where $k \in \{1, 2, \dots, K\} $. We introduce learnable scalar gating parameters $w_k$ for each adapter, which are transformed using a softmax function to ensure non-negative weights that sum to one:
$\alpha_k = \frac{\exp(w_k)}{\sum_{j=1}^K \exp(w_j)}$.
The combined representation $h_{\text{combined}} $ is computed as a weighted sum of adapter outputs:
$h_{\text{combined}} = \sum_{k=1}^K \alpha_k h_k$.

This dynamic gating mechanism is trained end-to-end with the rest of the model. During training, the learnable weights $w_k$ are updated to minimise the overall loss function, allowing the model to select the most informative adapters for each specific input. By adjusting these weights, the model can prioritise key adapters over others, effectively capturing the complex and evolving nature of troll behaviour.

\subsection{Rationale Selector}
Most LLMs based on decoder-only architectures, such as GPT-3 \cite{brown2020language}, have demonstrated strong capabilities in natural language understanding and generation.
To leverage these strengths for explainable troll detection, we introduce a unified decoder architecture that simultaneously performs rationale selection and user classification--a new way over conventional two-stage approaches.

Our approach uses a shared decoder that jointly extracts informative rationales from user posts and performs classification based on these rationales. This design tries to mitigate the degeneration issue prevalent in two-phase rationalisation models, where classifiers often overfit to uninformative rationales \cite{liu2022fr}. By establishing direct interaction between rationale extraction and classification, our model creates a reinforcing cycle: classification guides the selection of relevant evidence, while focused rationales improve classification accuracy.

Formally, given a user timeline $T_u = [x_{(u,1)}, \ldots, x_{(u,n)}]$, we extract token-level rationales $\mathcal{R}_u = \{r_{(u,1)}, \ldots, r_{(u,k)}\}$ that serve as supporting evidence for classification. Unlike post-level approaches, our token-level selection precisely identifies linguistic cues that contribute to troll detection, better enhancing interpretability.

The rationale selection process, detailed in \cref{alg:selector}, operates as follows. Given user timeline $T_u$, we first concatenate all posts and compute contextual embeddings for each token. For token position $i$, we compute an attention score:
$
p_\psi(r_{k=1} \mid x_u) = f_\psi(x_u)_k,
$ where $f_\psi$ is our rationale selector parameterised by $\psi$, and $f_\psi(x_u)_k$ represents the attention probability assigned to token $x_{u,k}$. We retain tokens with probability exceeding threshold $\tau$:
\begin{align*}
r_k &= \mathbf{I}[p_\psi(r_{k=1} \mid x_u) > \tau] \\
\mathcal{R}_u &= \{x_{(u,k)} \mid r_{k = 1}\},
\end{align*}
where $\mathbf{I}[\cdot]$ is the indicator function and $\tau = 0.5$.

Without explicit gold-standard annotations for rationales, we apply two regularisation techniques to ensure high-quality selection:
\paragraph{Sparsity Constraint} We limit selected tokens to a maximum of $l$ or fraction $\alpha$ of input length:
\begin{equation*}
\sum_{k=1}^{|x_u|} r_k \leq \min(l, \alpha|x_u|).
\end{equation*}
\paragraph{Continuity Regularization} We encourage selection of coherent linguistic spans by penalising discontinuities:
\begin{equation*}
\label{eq:conti_loss}
\mathcal{L}_\text{cont} = \sum_{k=2}^{|x_u|} |r_k - r_{k-1}|.
\end{equation*}
When enforcing the sparsity constraint (\cref{alg:selector}, lines 8-12), we implement a dynamic selection process that prioritizes tokens with highest attention scores within the constraint budget. The continuity regularization (line 14) adopts a dynamic programming approach to find optimal contiguous spans that minimize discontinuity loss while respecting sparsity constraints.

The selected rationales serve dual purposes: they provide interpretable evidence for model predictions and support the generation of explanations. For classification, we pool the embeddings of selected rationale tokens and compute classification logits (lines 16-18):
\begin{equation*}
y_u = \sigma(\mathbf{w}_c^T \text{Pool}(\{H_i \mid r_i = 1\}) + b_c),
\end{equation*}
where $\text{Pool}(\cdot)$ aggregates rationale token embeddings and $\sigma$ is the sigmoid activation function.

Our unified approach creates a cycle where better rationales lead to more accurate classification, which in turn guides more precise rationale selection. This self-reinforcing mechanism outperforms traditional pipeline approaches, as demonstrated in our experimental results (\cref{sec:experiments}).

\begin{algorithm}
\caption{Rationale Selection}
\begin{algorithmic}[1]
\Require User timeline $T_u = [x_{(u,1)}, x_{(u,2)}, ..., x_{(u,n)}]$, threshold $\tau$, sparsity constraint $\alpha$, continuity weight $\lambda_c$
\Ensure Rationales $\mathcal{R}_u$, troll classification $y_u$

\State \textbf{Initialize} attention scores $A = []$, rationale binary mask $r = []$
\State \textbf{Concatenate} timeline posts: $x_u = [x_{(u,1)}; x_{(u,2)}; ...; x_{(u,n)}]$
\State \textbf{Compute} contextual embeddings: $H = \text{Encoder}(x_u) \in \mathbb{R}^{|x_u| \times d}$
\For{each token position $i \in \{1,2,...,|x_u|\}$}
   \State Compute token-level attention score: $a_i = \sigma(\mathbf{w}_a^T H_i + b_a)$
   \State Append to attention scores: $A = A \cup \{a_i\}$
\EndFor
\State \textbf{Apply} threshold: $r_i = \mathbf{1}[a_i > \tau]$ for all $i$
\If{$\sum_i r_i > \alpha|x_u|$} 
   \State Sort attention scores: $A_{\text{sorted}} = \text{Sort}(A, \text{descending}=\text{True})$
   \State Keep top-$k$ tokens where $k = \lfloor \alpha|x_u| \rfloor$
   \State Set mask $r_i = 1$ for tokens with top-$k$ attention scores, else $0$
\EndIf
\State \textbf{Apply} continuity regularisation:
\State \quad Minimise $\mathcal{L}_{\text{cont}}$ as in \cref{eq:conti_loss}
\State \textbf{Extract} rationale tokens: $\mathcal{R}_u = \{x_{(u,i)} | r_{i = 1}\}$
\State \textbf{Compute} classification logits:
\State \quad $h_{\mathcal{R}} = \text{Pool}(\{H_i | r_{i = 1}\})$
\State \quad $y_u = \sigma(\mathbf{w}_c^T h_{\mathcal{R}} + b_c)$
\State \Return $\mathcal{R}_u, y_u$
\end{algorithmic}
\label{alg:selector}
\end{algorithm}

\subsection{Summary Generation}

The summary generator $s_\omega$ in X-Troll produces concise, natural language explanations derived from the selected rationales $\mathcal{R}$, offering clear justifications for classification decisions. Inspired by \cite{tennenholtz2024demystifying}, we propose a troll-specific summary generator that incorporates rationale embeddings to enhance explanation quality.

Given a selected rationale $r \in \mathcal{R}$, we construct an explanatory summary $S$ using a base model augmented with an adapter layer $E_A$:
\begin{equation*}
S = s_\omega(r) = \text{LLM}\left( [ \texttt{CLS} ] \oplus E_A(r) \oplus [ \texttt{RAT} ] \right),
\end{equation*}
where $[ \texttt{CLS} ] $ and $[ \texttt{RAT} ]$ are special tokens indicating the start of the sequence and the rationale segment, respectively, and $\oplus$ denotes concatenation. The adapter $E_A: \mathbb{R}^d \rightarrow \mathbb{R}^e $ maps the rationale embedding into the model's token embedding space, ensuring seamless integration of the rationale information into the language model.

The adapter $E_A$ is implemented as a two-layer multilayer perceptron (MLP):
\begin{equation*}
E_A(r) = W_2 \cdot \text{ReLU}(W_1 r + b_1) + b_2,
\end{equation*}
where $W_1 \in \mathbb{R}^{h \times d}$, $W_2 \in \mathbb{R}^{e \times h}$, $b_1 \in \mathbb{R}^h $, and $b_2 \in \mathbb{R}^e$ are learnable parameters. Here, $d$ is the dimension of the rationale embedding, $h$ is the hidden dimension of the MLP.


\section{Experiments}
\label{sec:experiments}
This section presents experimental findings across four areas: few-shot learning, ablation study, summary generation, and a qualitative case study. Few-shot (zero/one/five-shot) evaluations used a held-out test set, with LoRA adapters trained independently per task. We used AdamW (learning rate 1e-3, weight decay 0.01), training for 10 epochs with early stopping.

For k-shot evaluation, examples were randomly sampled while maintaining a balanced positive-negative distribution. A consistent test set was used across all shot settings for comparability. The experiments were run on PyTorch 2.0 using 4 NVIDIA A100 GPUs (40GB each). Results were averaged over five runs with different random seeds for robustness\footnote{To ensure fair evaluation, we used consistent data splits across all experiments. The annotated dataset was divided into train:validation:test splits with ratios of 70:10:20.}.



\begin{table}[!htbp]
\centering
\caption{Data Statistics. The ``appraisals'' column indicates the number of tweets annotated with appraisal labels in each campaign category. Categories without appraisal annotations are marked with a dash (--).}
\begin{tabular}{lccc}
\toprule
\toprule
\textbf{Campaign} & \textbf{\#users} & \textbf{\#posts} &\textbf{\#appraisals}  \\ \midrule
Russia-Anti-NATO Troll & 70 & 26,684  & 124 \\ \hline
Russia-Anti-NATO Non-Troll & 140 & 36,895  & -- \\ \hline
Russia-IRA Troll & 31 & 68,914 & 159   \\ \hline
Russia-IRA Non-Troll & 100 & 34,511 & -- \\ \hline
PRC-Xinjiang Troll & 257 & 24,075 & 303 \\ \hline
PRC-Xinjiang Non-Troll & 1,444 & 356,112 &--  \\ \hline
Random  & 2,000 & 40,000  & --\\ \hline
\bottomrule
\end{tabular}
\label{tab:data_statistics}
\end{table}

\subsection{Datasets}
For troll detection and information campaign classification tasks, we used datasets from three specific state-sponsored information campaigns: 
Russia-Anti-NATO Troll, Russia-IRA Troll, and PRC-Xinjiang Troll\footnote{\url{https://blog.x.com/en_us/topics/company/2021/disclosing-state-linked-information-operations-we-ve-removed}}. 
These datasets were released by Twitter and contain troll accounts and their posts. 
Twitter banned these accounts in October 2018 and anonymized them to ensure they could not be associated with individual users.
For example, the `Russia-Anti-NATO Troll' dataset contains users banned for amplifying narratives that sought to undermine faith in the NATO alliance and its stability\footnote{\url{https://blog.x.com/en_us/topics/company/2021/disclosing-networks-of-state-linked-information-operations-}}. 
We also collected posts from non-troll users via the Twitter data API to establish a baseline dataset for evaluating model performance on non-troll data.
In our non-troll data collection, we excluded all profile information and retrieved only necessary data. 
We focused on posts related to specific topics to protect user privacy. 
\cref{tab:data_statistics} presents the summary of the data statistics including the number of users and posts.
To enhance analysis quality, domain experts annotated a subset of posts with appraisal labels for each information campaign, as shown in \cref{tab:data_statistics}. 
The propaganda strategy adapter was fine-tuned on datasets provided by \citet{moral2023overview} as part of the DIPROMATS 2023 shared task\footnote{\url{https://sites.google.com/view/dipromats2023}}.
We also provide a sample of annotated posts with detailed labels (including appraisal and propaganda tags), which is included in Appendix~\cite{appendixOnline}.



\subsection{Base Models}
We evaluate X-Troll across diverse model architectures to demonstrate the generalizability of our linguistic knowledge integration approach. Our evaluation includes four base language models: three decoder-only architectures (LlaMA-3-8B~\cite{touvron2023llama}, Falcon-7B~\cite{almazrouei2023falcon}, Gemma-7B~\cite{team2024gemma}) and one encoder-decoder model (FLAN-T5-XL~\cite{chung2024scaling}).
We compare X-Troll against established baselines across multiple settings: in-context learning and LoRA fine-tuning for all base models, GPT-4~\cite{achiam2023gpt} in-context learning, and MetaTroll~\cite{tian2023metatroll}, the current state-of-the-art BERT-based few-shot troll detection system. 


\begin{table}[t]
\caption{Model performance (F$_1$)on troll detection and campaign classification tasks under few-shot settings. Results are reported for three settings: 0S (zero-shot), 1S (one-shot), 5S (five-shot).}
\label{tab:few-shot}
\centering
\small
\begin{tabular}{lcccccc}
\toprule
\toprule
\multirow{2}{*}{Model} & \multicolumn{3}{c}{Troll Detection} & \multicolumn{3}{c}{Campaign Classification} \\
& 0S & 1S & 5S & 0S & 1S & 5S \\
\midrule
\multicolumn{7}{l}{\textbf{In-context learning}} \\
LLaMA & 0.456 & 0.502 & 0.549 & 0.328 & 0.375 & 0.421 \\
Falcon & 0.487 & 0.535 & 0.582 & 0.359 & 0.407 & 0.456 \\
Gemma & 0.505 & 0.553 & 0.600 & 0.377 & 0.425 & 0.474 \\
T5 & 0.421 & 0.468 & 0.515 & 0.293 & 0.339 & 0.386 \\
GPT-4 & \textbf{0.523} & 0.571 & 0.618 & \textbf{0.396} & \textbf{0.444} & \textbf{0.493} \\
MetaTroll & - & \textbf{0.582} & \textbf{0.689} & - & - & - \\
\midrule
\multicolumn{7}{l}{\textbf{LoRA fine-tuned}} \\
LLaMA & 0.535 & 0.580 & 0.630 & 0.390 & 0.435 & 0.485 \\
Falcon & 0.570 & 0.615 & 0.665 & 0.425 & 0.470 & 0.520 \\
Gemma & \textbf{0.585} & \textbf{0.630} & \textbf{0.680} & \textbf{0.440} & \textbf{0.485} & \textbf{0.535} \\
T5 & 0.500 & 0.545 & 0.595 & 0.355 & 0.400 & 0.450 \\
\midrule
\multicolumn{7}{l}{\textbf{X-Troll}} \\
LLaMA & 0.592 & 0.628 & 0.663 & 0.483 & 0.518 & 0.552 \\
Falcon & 0.627 & 0.661 & 0.696 & 0.522 & 0.557 & 0.591 \\
Gemma & \textbf{0.648} & \textbf{0.682} & \textbf{0.717} & \textbf{0.547} & \textbf{0.581} & \textbf{0.616} \\
T5 & 0.558 & 0.593 & 0.627 & 0.448 & 0.482 & 0.516 \\
\bottomrule
\bottomrule
\end{tabular}
\end{table}

\subsection{Detection Performance}
X-Troll show strong performance over both general-purpose LLMs and specialized transformer models across all evaluation settings. \cref{tab:few-shot} shows that X-Troll (Gemma-7B) achieves F$_1$ scores of $0.648$, $0.682$, and $0.717$ for zero-shot, one-shot, and five-shot troll detection respectively--consistent $6.3$, $5.2$, and $3.7$ percentage point improvements over the best baseline (LoRA fine-tuned Gemma-7B).

\noindent\textbf{The Expert Knowledge Advantage.} The performance hierarchy reveals critical insights about automated troll detection capabilities. While general-purpose LLMs struggle with subtle manipulation techniques (GPT-4: $0.523$ zero-shot F$_1$), specialized fine-tuning provides improvements (LoRA Gemma-7B: $0.585$). However, X-Troll's systematic integration of linguistic expertise ($0.648$) shows that expert knowledge can bridge remaining performance gaps--particularly crucial given the evolving sophistication of state-sponsored operations.

\noindent\textbf{Campaign Classification Insights.} For information campaign classification, the linguistic advantage becomes even more pronounced. X-Troll achieves $10.7$, $9.6$, and $8.1$ percentage point improvements over LoRA baselines across few-shot settings, suggesting that campaign-specific linguistic signatures are particularly distinctive and can be effectively captured through our multi-adapter architecture. This finding validates our hypothesis that different state-sponsored operations use systematically different rhetorical strategies.

\noindent\textbf{Practical Deployment Implications.} The performance gap between zero-shot ($0.648$ F$_1$) and five-shot ($0.717$ F$_1$) settings shows that X-Troll can provide reasonable detection capabilities even with minimal training examples from new campaign types. This rapid adaptation capability is essential for countering emerging threats where extensive labeled data may not be immediately available, demonstrating how linguistic expertise enables practical deployment in operational environments where traditional approaches would require extensive retraining.

\begin{figure}[t]
  \centering
  \includegraphics[width=\linewidth]{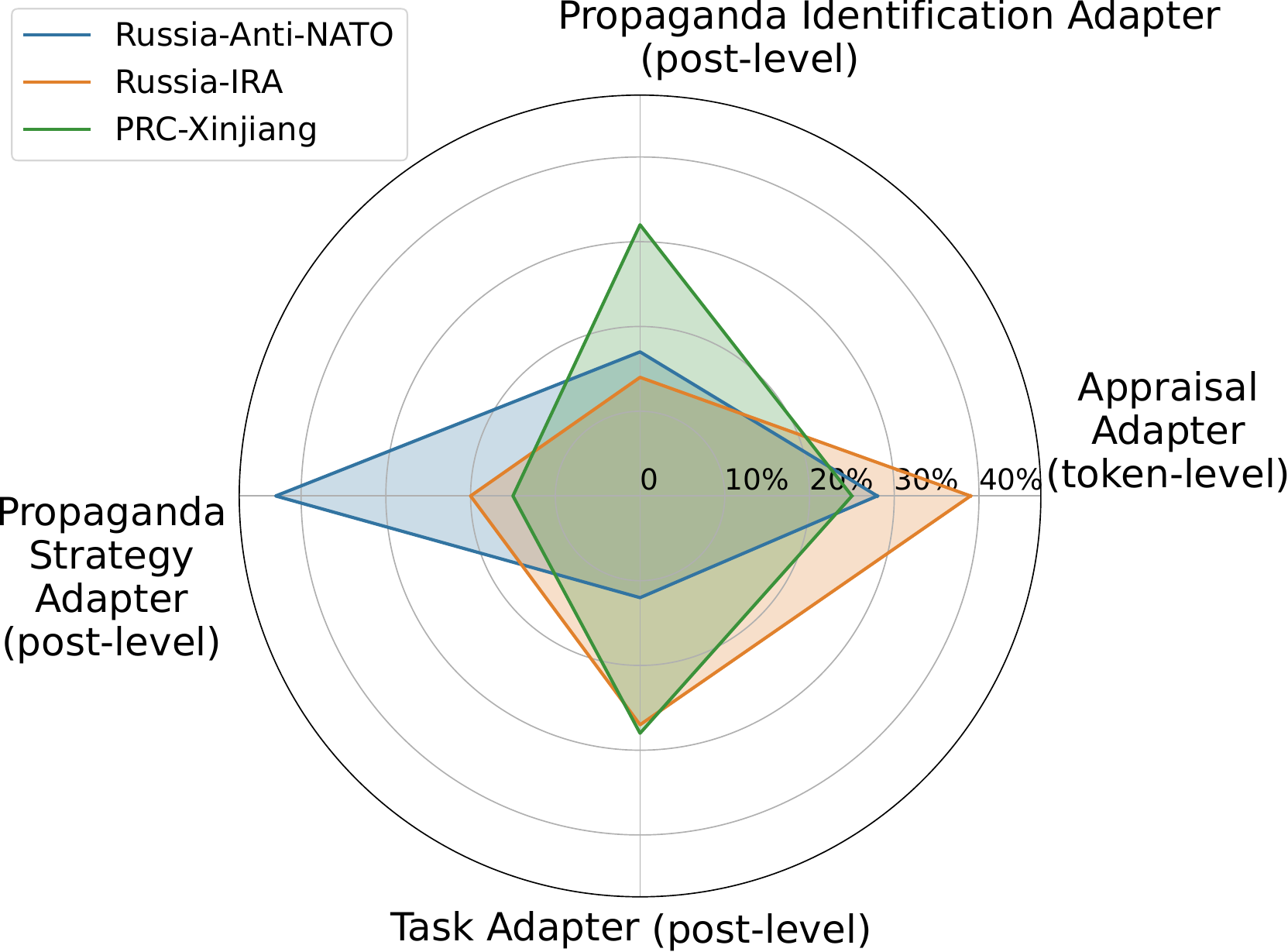}
  \caption{Adapter weight distribution across information operations. The radar chart illustrates the relative weighting of four adapter types (Appraisal, Propaganda Identification, Propaganda Strategy, and Task) across three different information operations (Russia-Anti-NATO (blue), Russia-IRA (orange), and PRC-Xinjiang (green)).}
  \label{fig:gating}
\end{figure}
\subsection{Campaign-Specific Patterns}
Our adapter weight analysis uncovers distinct behavioral patterns across state-sponsored information operations, providing unprecedented insights into the strategic doctrines underlying different campaigns. \cref{fig:gating} reveals how different operations exhibit unique linguistic signatures that align with documented strategic approaches.

\noindent\textbf{Russia-Anti-NATO: Strategy-Driven Sophistication.} Russia-Anti-NATO campaigns show strong reliance on the Propaganda Strategy adapter ($0.43$) while minimizing direct Task-specific ($0.12$) and Propaganda Identification ($0.17$) features. This pattern reflects sophisticated rhetorical approaches that prioritize subtle strategic techniques over overt propaganda signals. The emphasis on nuanced strategy aligns with documented ``gray zone'' warfare approaches characteristic of Russian information operations, where plausible deniability requires detailed rhetorical sophistication~\cite{mazarr2015mastering,barrinha2017cyber}.

\noindent\textbf{Russia-IRA: Appraisal-Focused Narrative Construction.} In contrast, Russia-IRA operations prioritize the Appraisal adapter ($0.39$) and Task features ($0.27$) while showing reduced dependence on Propaganda Strategy ($0.20$). This pattern validates findings by ~\citet{lazer2018science} documenting how Russian disinformation campaigns by IRA use evaluative language to construct persuasive narrative frames. The focus on appraisal-based manipulation suggests these operations emphasize subtle sentiment steering over explicit rhetorical techniques--a finding with strong implications for detection system design.

\noindent\textbf{PRC-Xinjiang: Balanced Multi-Modal Doctrine.} PRC-Xinjiang campaigns show a more balanced activation pattern with heightened Propaganda Identification ($0.32$) alongside maintained Appraisal ($0.25$) and Task-specific ($0.28$) weights. This balanced approach suggests a different operational doctrine that combines explicit propaganda techniques with subtle linguistic manipulation, possibly reflecting distinct cultural and strategic contexts for Chinese information operations compared to Russian approaches.

\noindent\textbf{Detection System Implications.} These findings provide three insights for troll detection task. First, they empirically validate that state-sponsored operations use campaign-specific linguistic strategies that can be systematically distinguished through computational analysis. Second, they demonstrate the necessity of multi-faceted detection approaches--relying solely on propaganda identification would systematically miss the appraisal-based strategies prevalent in Russian campaigns. Third, they validate our dynamic gating mechanism's ability to automatically focus on contextually relevant linguistic features without manual reconfiguration.
\begin{figure}[t]
\centering
\includegraphics[width=\linewidth]{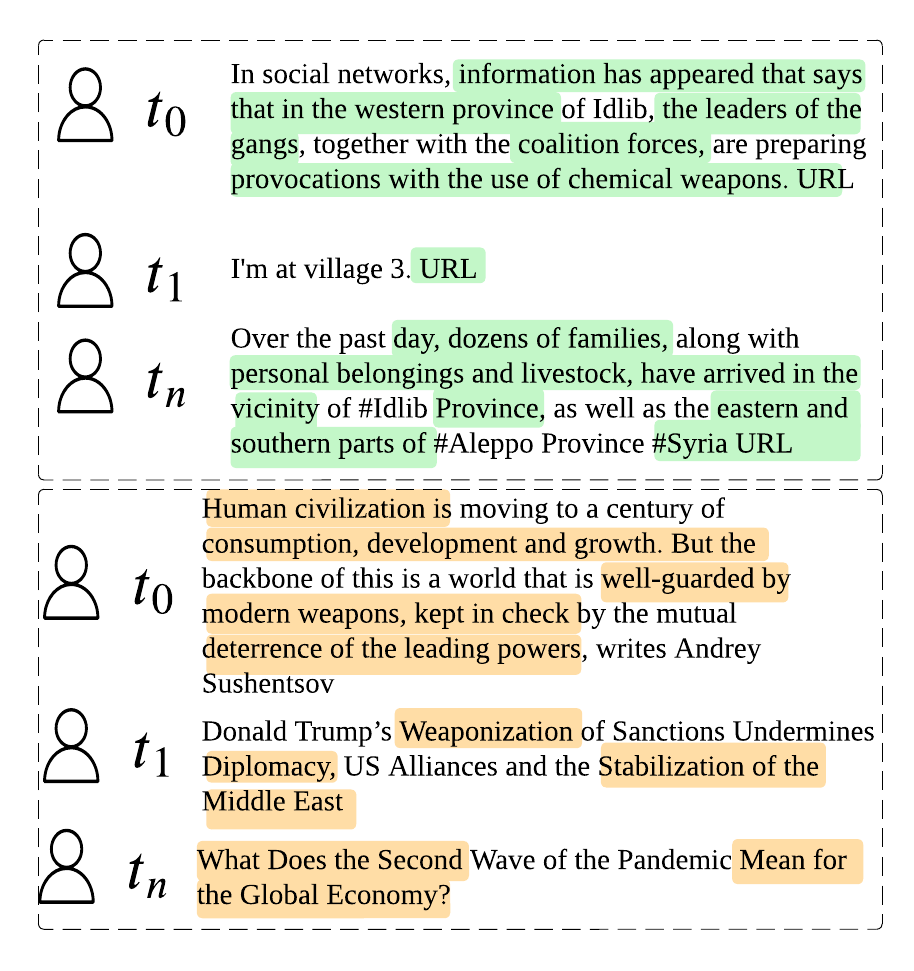}
\caption{X-Troll's rationale selection on Russia-IRA examples, with diagnostic tokens highlighted. The correctly classified troll post (top) shows characteristic geopolitical framing and conflict narratives, while the false positive (bottom) reveals how political topic overlap without coordinated rhetorical patterns can mislead classification.}
\label{fig:case_study}
\end{figure}

\subsection{Explainability Analysis}
We analyze X-Troll's explainability through three dimensions: summary generation evaluation, case analysis of successful rationale selection and error analysis of false positives.

\noindent\textbf{Summary Generation Evaluation.} \cref{tab:summary-generator} reveals certain patterns in explanation generation performance. The rationale selector is able to improve explanation quality across most model-dataset combinations, with strong improvements for FLAN-T5 on Russia-IRA data ($7.1$\% improvement) and Falcon-7B on PRC-Xinjiang data ($6.2$\% improvement). However, some configurations show slight degradation (FLAN-T5 on Russia-Anti-NATO: $-4.5$\%), suggesting that rationale effectiveness varies with campaign characteristics and base model capabilities.
In Appendix~\cite{appendixOnline}, we include the detailed G-Eval prompts and templates used for scoring~\citep{liu2023g}, as well as the complete evaluation results, reporting the scores for each evaluation criterion.

\noindent\textbf{Mechanistic Insights from Case Analysis.} \cref{fig:case_study} illustrates both the power and limitations of our rationale-based approach. In correctly identified cases, X-Troll successfully highlights emotionally charged phrases (``leaders of the gangs,'' ``provocations'') that align with established propaganda techniques--opponent discrediting and loaded language respectively. This demonstrates  contextual pattern recognition beyond simple emotional language detection, validating the value of token-level evidence extraction over post-level approaches.

\noindent\textbf{False Positive Pattern Analysis.}
Misclassification analysis reveals the challenges in distinguishing trolling patterns from legitimate political discourse. \cref{fig:case_study} (bottom) shows a representative false positive where X-Troll incorrectly classified an authentic user as a troll.
In this case, the system focuses on ideological framing language (``human civilization'') and geopolitical discourse markers rather than manipulation-specific linguistic patterns characteristic of coordinated information operations. The rationale selector highlights phrases that reflect legitimate geopolitical analysis rather than the systematic opponent discrediting or loaded language patterns typical of state-sponsored trolls. 
This error shows that the system can be misled by complex political discourse that uses abstract conceptual framing without manipulative intent.

\begin{table}[t]
\centering
\caption{Impact of rationale selection on explanation quality across three information campaigns. Scores are G-Eval ratings (1-5 scale) for coherence, consistency, fluency, and relevance. 
Percentage changes indicate relative improvement (+) or degradation (-) from baseline. \textbf{Bold} indicates stongest improvements per campaign; \textit{italics} denote performance decreases.}
\resizebox{\columnwidth}{!}{%
\begin{tabular}{lccc}
\toprule
\toprule
\textbf{Model} & \textbf{Russia-Anti-NATO} & \textbf{Russia-IRA} & \textbf{PRC-Xinjiang} \\
\midrule
\multicolumn{4}{c}{X-Troll w/o rationale selector} \\
\midrule
LLaMA-3B & \textbf{3.607} & 3.231 & 3.060 \\
Falcon-7B & 3.032 & \textbf{3.348} & 3.292 \\
Gemma-7B & 3.482 & 3.127 & \textbf{3.524} \\
FLAN-T5 & 2.904 & 2.609 & 2.937 \\
\midrule
\multicolumn{4}{c}{X-Troll with rationale selector} \\
\midrule
LLaMA-3B & \textbf{3.847} (\textbf{+6.7\%}) & \textit{3.186} (\textit{-1.4\%}) & 3.075 (+0.5\%) \\
Falcon-7B & 3.148 (+3.8\%) & 3.474 (+3.8\%) & \textbf{3.496} (\textbf{+6.2\%}) \\
Gemma-7B & 3.552 (+2.0\%) & 3.256 (+4.1\%) & 3.536 (+0.3\%) \\
FLAN-T5 & \textit{2.774} (\textit{-4.5\%}) & \textbf{2.796 }(\textbf{+7.1\%}) & 3.059 (+4.1\%) \\
\bottomrule
\bottomrule
\end{tabular}%
}
\label{tab:summary-generator}
\end{table}

\subsection{Ablation Studies}
\begin{figure}[t]
  \centering
  \includegraphics[width=\linewidth]{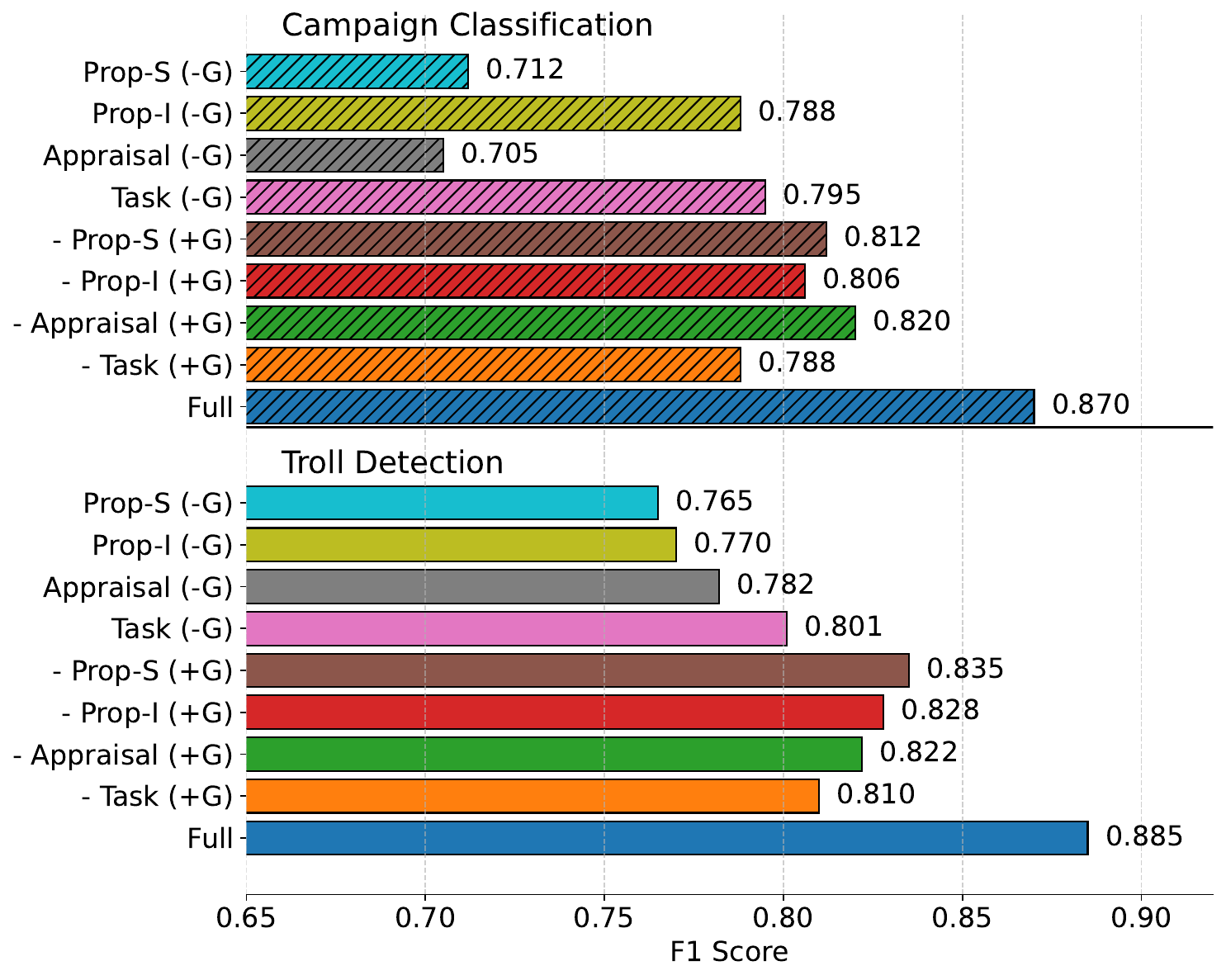}
  \caption{Ablation study results showing F1 scores for troll detection and campaign classification tasks. Each bar represents a configuration: ``Full'' uses all four adapters with gating, ``- [Adapter] (+G)'' removes one adapter while keeping the gating mechanism, and ``[Adapter] (-G)'' uses only a single adapter without gating. Prop-I = Propaganda Identification adapter, Prop-S = Propaganda Strategy adapter.}
  \label{fig:ablation}
\end{figure}

To evaluate the impact of individual adapters in X-Troll, we conducted ablation studies shown in \cref{fig:ablation}. 
We examine two configurations to quantify the role of adapters and impact of expert knowledge integration. 
First, we remove individual adapters while retaining the gating mechanism (+G). 
Second, we test minimal configurations where only one adapter operates without gating (-G). 

The full X-Troll model achieves the highest F$_1$ scores of $0.885$ for troll detection and $0.870$ for campaign classification. 
This confirms that adapter fusion and expert knowledge integration significantly enhance classification accuracy. 
Performance degradation upon adapter removal reveals a hierarchical importance among knowledge sources.

\noindent\textbf{Task Adapter Centrality.} \cref{fig:ablation} demonstrates that the Task Adapter proves most critical for both troll detection ($-7.5$\% when removed) and campaign classification ($-8.2$\%). This finding validates the importance of direct supervised learning from troll-labeled data while highlighting how expert knowledge augments rather than replaces task-specific adaptation.

\noindent\textbf{Complementary Expert Knowledge Effects.} Appraisal and Propaganda adapters show comparable importance levels ($5-6$\% performance drops when removed), but minimal configurations reveal their interdependence. Single adapters alone show performance degradation (up to $16.5$\% for campaign classification with Appraisal only), showing that expert knowledge sources achieve maximum effectiveness through integration rather than isolation.

\noindent\textbf{Dynamic Gating Validation.} The comparison between gated (+G) and non-gated (-G) configurations validates our dynamic integration approach. Without gating, the Task Adapter alone achieves only $0.801$ F$_1$ for troll detection and $0.795$ for campaign classification, representing substantial degradation from the full model's $0.885$/$0.870$ performance. 
In contrast, non-gated single adapters suffer dramatic performance losses (Appraisal alone: $0.782$/$0.705$ F$_1$, $-11.6$\%/$-19.0$\% drops), demonstrating strong complementarity effects. The performance gap between gated and non-gated configurations (e.g.,\ Appraisal: $0.822$ vs. $0.782$) confirms that effective troll detection requires dynamic integration of multiple knowledge sources rather than static combination approaches.

\section{Conclusion}
We introduced X-Troll, an explainable framework that integrates linguistic expert knowledge with large language models for state-sponsored troll detection. By systematically incorporating appraisal theory and propaganda analysis through specialized LoRA adapters, X-Troll is trying to address the limitations in current detection systems: poor interpretability and inability to capture sophisticated manipulation strategies.
Our evaluation demonstrates that linguistic knowledge integration provides substantial benefits, achieving $5-10$ percentage point improvements over strong baselines while generating human-readable explanations. 
The dynamic gating mechanism reveals distinct strategic patterns across information operations, with Russian campaigns emphasizing appraisal-based manipulation while Chinese operations use more balanced propaganda techniques.
Key findings extend beyond detection performance. Our hierarchical knowledge integration shows that expert linguistic insights achieve maximum effectiveness when combined with task-specific learning, providing a template for incorporating domain expertise in security applications. 

X-Troll's explainable approach contributes to a growing ecosystem of tools for understanding and countering online manipulation. While \citet{Kong2023} focuses on detecting operations through social reactions and \citet{Tian2025ICMamba} enables early prediction of content engagement, X-Troll provides the crucial missing piece: explainable identification of the actors themselves. Future work could integrate these complementary approaches, combining early engagement prediction with explainable actor detection to create comprehensive early warning systems for information operations.

\section*{Acknowledgements}
This research was supported by the Australian Research Council Discovery Project DP200101441, the Advanced Strategic Capabilities Accelerator (ASCA), the Australian Department of Home Affairs, Commonwealth of Australia as represented by the Defence Science and Technology Group of the Department of Defence, and the Defence Innovation Network.

\newpage
\section*{GenAI Usage Disclosure}
This work was created, reviewed, and edited by human authors. AI tools were used in two specific capacities: (1) debugging the code components of the X-Troll framework, and (2) writing assistance to improve conciseness and readability of manuscript sections.

For writing assistance, we used Claude (Anthropic) with prompts such as: ``Improve the writing of this paragraph from a scientific paper. Keep concise, and improve reading flow. Match style. Highlight changes. Break down complex and long sentences and make more concise.'' All AI-generated suggestions were critically reviewed, modified, and integrated by human authors. The original conceptual content, technical contributions, experimental design, analysis, and final editorial decisions remain entirely human-authored. AI tools did not contribute to the research methodology, data analysis, or scientific conclusions.

\bibliographystyle{ACM-Reference-Format}
\bibliography{ref}

\input{appendix}

\end{document}

%% file: appendix.tex
\balance  
\newpage
\appendix
\section*{Appendix}

\section{Sample Annotated Data}
\begin{figure*}[t]
\centering
  \includegraphics[width=\linewidth]{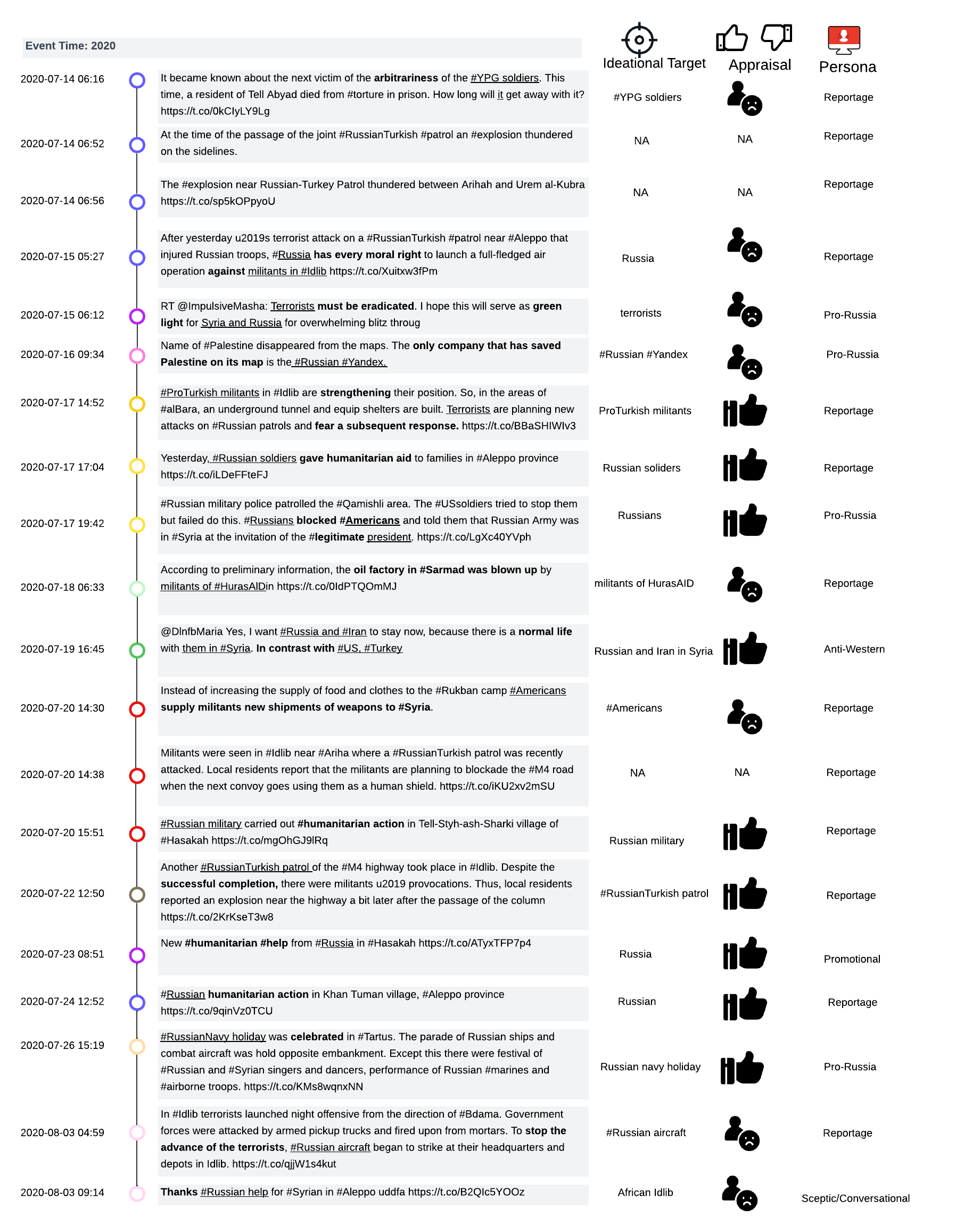}
  \caption{Sample of annotated posts from the Russia–Syria dataset, showing text excerpts with corresponding expert labels (ideational target, persona, appraisal, and propaganda technique). Full datasets are described in Section 5.1.}
  \label{fig:sample_data}
\end{figure*}

We provide a sample of our annotated data with detailed labels (see~\cref{fig:sample_data}).
While the main experiments rely on large-scale datasets described in Section 5.1, this illustrative sample shows our annotation schema and the kinds of linguistic patterns X-Troll is designed to capture.

State-sponsored troll detection requires moving beyond post-level classification to understand \textit{how} manipulation strategies unfold linguistically across different narrative contexts. Our annotation framework captures the complex interplay between ideational targeting, evaluative stance, and persona construction that characterizes coordinated inauthentic behavior. 

Our annotation framework captures three interconnected dimensions of state-sponsored discourse. The ideational target dimension identifies which entities—from specific groups—are being framed in each post, revealing strategic attention direction. The appraisal dimension encodes evaluative stance toward these targets, capturing subtle positioning shifts that traditional sentiment analysis might miss. The persona dimension characterizes the discursive role adopted by accounts, showing how agents modulate voice for credibility and detection evasion.

The sample shows 2020 Syrian conflict discourse, where these dimensions interact to produce sophisticated manipulation. Posts combining ``reportage'' personas with selective military targeting exemplify how neutral news-sharing masks geopolitical agenda-setting.

\section{Summary Evaluation}
\label{sec:summ_eval_appendix}
\begin{table*}[t]
\centering
\caption{Generated summary evaluation on detailed metrics using Gemma-7B as the base model.}
\label{tab:summ_eval_gemma}

\begin{tabular}{lccccccc}
\toprule
\toprule
& \multicolumn{2}{c}{\textbf{Russa-Anti-NATO}} & \multicolumn{2}{c}{\textbf{Russia-IRA}} & \multicolumn{2}{c}{\textbf{PRC-Xinjiang}} \\
\cmidrule(lr){2-3} \cmidrule(lr){4-5} \cmidrule(lr){6-7}
\textbf{Metric} & \textbf{w/o rational} & \textbf{X-Troll} & \textbf{w/o rational} & \textbf{X-Troll} & \textbf{w/o rational} & \textbf{X-Troll} \\
\midrule
Coherence & 2.25 & 2.60 & 2.15 & 2.65 & 2.60 & \textbf{2.90} \\
Consistency & 3.50 & 3.35 & 3.25 & 3.30 & 3.50 & 3.50 \\
Fluency & 4.90 & 4.95 & 4.95 & 4.90 & 4.90 & 4.95 \\
Relevance & 3.40 & \textbf{3.55} & 2.25 & 2.30 & 3.35 & 2.95 \\
\bottomrule
\bottomrule
\end{tabular}

\end{table*}
We evaluated our approach across three influence campaigns (Russia-Anti-NATO, Russia-IRA, and PRC-Xinjiang) using four metrics: coherence, consistency, fluency, and relevance. \cref{tab:summ_eval_gemma} compares X-Troll with and without the rationale selector using Gemma-7B as the base model.

Fluency consistently scored around $5.0$ across all datasets, indicating that both models maintain high linguistic quality. Coherence showed the most significant challenge, with scores between 2.0 and 3.0, highlighting potential areas for improvement. Consistency and relevance exhibited moderate scores ($3.0$--$3.5$) across campaigns.

While X-Troll improved coherence and consistency in most cases, its impact on relevance varied, suggesting potential over-filtering effects. The minimal performance gap across metrics suggests that X-Troll preserves baseline strengths while enhancing interpretability, demonstrating robustness across different campaign contexts.


\section{Summary Evaluation Prompt Template}
\label{sec:prompts}

This section shows the evaluation prompts used to assess the quality generated summaries in X-troll using G-Eval score~\citep{liu2023g}. Each prompt includes detailed instructions, evaluation criteria, and a scoring guide.
The \texttt{[BACKGROUND]} placeholder in the prompt is filled with relevant information about the specific information campaigns from our datasets, providing context for the evaluation.
We evaluate each generated summary individually.
All four aspects (coherence, consistency, fluency and relevance) are scored on a scale of 1-5.

\begin{tcolorbox}[title=Prompt template for Coherence]
\ttfamily
\textbf{System:} You will be given one summary written for summarising a user's posts in the timeline.\\

Your task is to rate the summary on one metric.\\

Please make sure you read and understand these instructions carefully. Please keep this document open while reviewing, and refer to it as needed.\\

\textbf{Evaluation Criterion:}\\
Coherence (1-5) - the collective quality of all sentences. We align this dimension with the DUC quality question of structure and coherence whereby "the summary should be well-structured and well-organized. The summary should not just be a heap of related information, but should build from sentence to a coherent body of information about the topic. Here is the background information about this information campaign [BACKGROUND]."\\

\textbf{Evaluation Steps:}
\begin{enumerate}[leftmargin=1.5em, topsep=0pt, itemsep=2pt]
\item Read the news article carefully and identify the main topic and key points.
\item Read the summary and compare it to the source posts. Check if the summary covers the main topic and key points of all the source posts, and if it presents them in a clear and logical order.
\item Assign a score for coherence on a scale of 1 to 5, where 1 is the lowest and 5 is the highest based on the Evaluation Criteria.
\end{enumerate}

\textbf{Evaluation Form (score ONLY):}\\
- Coherence:
\end{tcolorbox}

\begin{tcolorbox}[title=Prompt template for Consistency]
\ttfamily
\textbf{System:} You will be given a timeline with source posts from a user. You will then be given one summary written for this user.\\

Your task is to rate the summary on one metric.\\

Please make sure you read and understand these instructions carefully. Please keep this document open while reviewing, and refer to it as needed.\\

\textbf{Evaluation Criterion:}\\
Consistency (1-5) - the factual alignment between the summary and the source posts. A factually consistent summary contains only statements that are entailed by the source posts. Annotators are also asked to penalize summaries that contained hallucinated facts. Here is the background information about this information campaign [BACKGROUND].\\

\textbf{Evaluation Steps:}
\begin{enumerate}[leftmargin=1.5em, topsep=0pt, itemsep=2pt]
\item Read the source posts carefully and identify the main facts and details they present.
\item Read the summary and compare it to the source posts. Check if the summary contains any factual errors that are not supported by the source posts.
\item Assign a score for consistency based on the Evaluation Criteria.
\end{enumerate}

\textbf{Evaluation Form (score ONLY):}\\
- Consistency:
\end{tcolorbox}

\begin{tcolorbox}[title=Prompt template for Fluency]
\ttfamily
\textbf{System:} You will be given one summary written for a user's posts in the timeline.\\

Your task is to rate the summary on one metric.\\

Please make sure you read and understand these instructions carefully. Please keep this document open while reviewing, and refer to it as needed.\\

\textbf{Evaluation Criterion:}\\
Fluency (1-5): the quality of the summary in terms of grammar, spelling, punctuation, word choice, and sentence structure.

\begin{itemize}[leftmargin=1.5em, topsep=0pt, itemsep=2pt]
\item[1:] Very Poor. The summary has numerous severe errors that make it very difficult to understand or sound extremely unnatural.
\item[2:] Poor. The summary has many errors that significantly impact clarity and readability.
\item[3:] Fair. The summary has some errors that affect the clarity or smoothness of the text, but the main points are still comprehensible.
\item[4:] Good. The summary has few errors and is generally easy to read and follow.
\item[5:] Excellent. The summary has virtually no errors and reads very smoothly and naturally.
\end{itemize}

\textbf{Evaluation Form (score ONLY):}\\
- Fluency (1-5):
\end{tcolorbox}

\begin{tcolorbox}[title=Prompt template for Relevance]
\ttfamily
\textbf{System:} You will be given one summary written for a user's posts in the timeline.\\

Your task is to rate the summary on one metric.\\

Please make sure you read and understand these instructions carefully. Please keep this document open while reviewing, and refer to it as needed. Here is the background information about this information campaign [BACKGROUND].\\

\textbf{Evaluation Criteria:}\\
Relevance (1-5) - selection of important content from the source. The summary should include only important information from the source posts. Annotators are instructed to penalize summaries which contained redundancies and excess information.\\

\textbf{Evaluation Steps:}
\begin{enumerate}[leftmargin=1.5em, topsep=0pt, itemsep=2pt]
\item Read the summary and the source posts carefully.
\item Compare the summary to the source posts and identify the main points of the source posts.
\item Assess how well the summary covers the main points of the posts, and how much irrelevant or redundant information it contains.
\item Assign a relevance score from 1 to 5.
\end{enumerate}

\textbf{Evaluation Form (score ONLY):}\\
- Relevance:
\end{tcolorbox}




